\pdfoutput=1

\documentclass{article}

\usepackage[preprint]{neurips_2023}

\usepackage{amssymb}
\usepackage{marvosym}

\usepackage[utf8]{inputenc} 
\usepackage[T1]{fontenc}    
\usepackage{hyperref}       
\usepackage{url}            
\usepackage{booktabs}       
\usepackage{amsfonts}       
\usepackage{nicefrac}       
\usepackage{microtype}      
\usepackage{xcolor}         
\usepackage{multirow,multicol}
\usepackage{graphicx}
\usepackage{enumitem}
\usepackage{amsmath}
\usepackage{subcaption}
\usepackage{float}
\usepackage{listings}
\usepackage{tcolorbox}
\usepackage{tikz}
\usepackage{geometry}
\usepackage{rotating}

\title{AI-native Memory 2.0: Second Me}

\author{
  Jiale Wei$\quad$ Xiang Ying$\thanks{Project Lead}\quad$ Tao Gao$\quad$ Fangyi Bao$\quad$ Felix Tao$\quad$ Jingbo Shang\\
  \\
  \{yingxiang, tao\}@mindverse.ai \\
  \\
  Mindverse.ai
}

\usepackage{xspace}

\newcommand{\our}{\textsc{Second Me}\xspace}

\newcommand{\smallsection}[1]{\noindent\textbf{#1}.}

\begin{document}

\maketitle

\begin{abstract}
    Human interaction with the external world fundamentally involves the exchange of personal memory, whether with other individuals, websites, applications, or, in the future, AI agents. 
A significant portion of this interaction is redundant, requiring users to repeatedly provide the same information across different contexts. 
Existing solutions, such as browser-stored credentials, autofill mechanisms, and unified authentication systems, have aimed to mitigate this redundancy by serving as intermediaries that store and retrieve commonly used user data.  
The advent of large language models (LLMs) presents an opportunity to redefine memory management through an AI-native paradigm: \textbf{\our}. 
\our acts as an intelligent, persistent memory offload system that retains, organizes, and dynamically utilizes user-specific knowledge. 
By serving as an intermediary in user interactions, it can autonomously generate context-aware responses, prefill required information, and facilitate seamless communication with external systems, significantly reducing cognitive load and interaction friction.  
Unlike traditional memory storage solutions, \our extends beyond static data retention by leveraging LLM-based memory parameterization. 
This enables structured organization, contextual reasoning, and adaptive knowledge retrieval, facilitating a more systematic and intelligent approach to memory management. 
As AI-driven personal agents like \our become increasingly integrated into digital ecosystems, \our further represents a critical step toward augmenting human-world interaction with persistent, contextually aware, and self-optimizing memory systems.
We have open-sourced the fully localizable deployment system at GitHub: \url{https://github.com/Mindverse/Second-Me}. 
\end{abstract}

\section{Introduction}

Human interaction with the external world relies heavily on memory, whether recalling information in conversations or repeatedly providing personal data across digital platforms. 
This redundancy leads to cognitive fatigue and disrupts seamless engagement with technology. 
Existing solutions, such as autofill mechanisms and unified authentication systems, help alleviate some of this burden by storing and retrieving commonly used data. 
However, they function as static repositories without contextual reasoning or adaptability, requiring users to manually manage and verify their information, resulting in a fragmented and suboptimal experience.

\begin{figure}[t]
    \centering
    \includegraphics[width=0.8\linewidth]{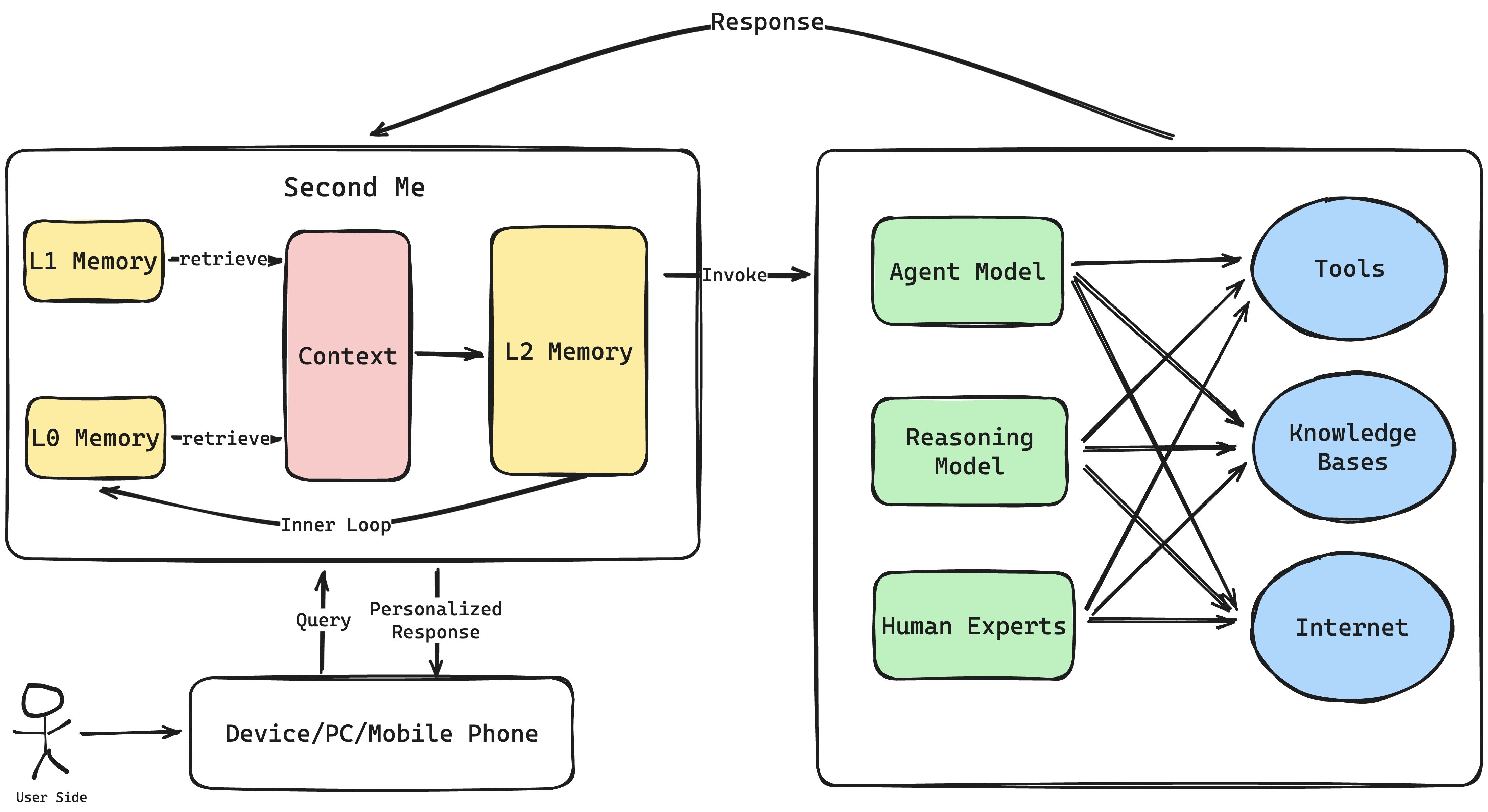}
    \caption{Hybrid Architecture of \our}
    \label{fig:1}
\end{figure}

The rise of large language models (LLMs) offers a transformative opportunity to redefine memory management through an AI-native approach. As shown in Figure~\ref{fig:1}, we introduce \our, an intelligent, persistent memory offload system that serves as a dynamic intermediary in human-machine interactions. Unlike traditional storage solutions, \our is an adaptive, context-aware assistant that autonomously retrieves, organizes, and applies user-specific knowledge. Leveraging LLM-based memory parameterization as envisioned in~\citet{shang2024ai}, \our enhances structured knowledge organization, contextual reasoning, and adaptive retrieval for seamless digital interactions.

In this work, we explore diverse data sources and training styles, integrating supervised fine-tuning (SFT) and direct preference optimization (DPO)~\citep{rafailov2024directpreferenceoptimizationlanguage} to enhance LLM performance.
We introduce three key tasks with automated LLM evaluation to assess the model's effectiveness in personal AI applications: (1) memory-based multi-perspective Q\&A, (2) context completion based on user needs, and (3) context critique incorporating user preferences and external responses.
To support these tasks, we design an LLM-driven automated data synthesis strategy that integrates local and global data perspectives, a multi-agent framework, and Chain-of-Thought (CoT)~\citep{wei2023chainofthoughtpromptingelicitsreasoning} style synthesis.
Our experiments show that using diverse data sources with strong CoT-style normalization yields the best \our performance in automated evaluations. 
Additionally, human case studies suggest that \our's true effectiveness may surpass reported metrics, as LLM-based evaluations tend to underestimate its quality.

To the best of our knowledge, we propose the first fully automated post-training pipeline based on personal document records, using the trained personal model as the core of a multi-layer hybrid system. 
This \our system would serve as a new paradigm for future personalized LLM applications.  
We have open-sourced the fully localizable deployment system at GitHub: \url{https://github.com/Mindverse/Second-Me}. 

Given \our's strong performance, we anticipate a vast range of applications in the near future. As an extension of human memory, \our reduces cognitive effort by preemptively providing relevant information, auto-completing forms, recalling past interactions, and maintaining context across applications. Beyond mere storage, it functions as a self-optimizing AI agent that learns, adapts, and refines its understanding of user preferences and behaviors.  

As AI-driven personal memory agents like \our integrate into digital ecosystems, they usher in a new era of human-AI collaboration. By enhancing memory retention and retrieval in a contextually aware manner, \our streamlines digital interactions, reduces friction, and improves efficiency, enabling a more seamless and intelligent exchange of knowledge. Furthermore, \our has the potential to facilitate \emph{networked intelligence}, where multiple agents collaborate, share relevant insights, and coordinate tasks across different applications and users. 

\section{An Overview of \our}

We envision \our as a Context Provider.
It acts as a bridge between users, future AI agents, and the broader information world, facilitating seamless interaction.
It is an evolved version of the Large Personal Model (LPM)~\citep{shang2024ai}, which was proposed by us last year, a three-layer system centered around personalized memory management. 
From a technical perspective, \our has further refined the Hybrid framework design and validated the iteration of personalized models, ensuring the system's efficiency and adaptability in addressing complex demands.
\our not only represents the cutting edge of our current technology but also embodies our product philosophy at this stage --- reflecting our commitment to personalized and intelligent interactions.

\subsection{Large Personal Model (LPM) 1.0: A Recap}

As we argued in~\citet{shang2024ai}, AI-native memory is a must-have component towards Artificial General Intelligence (AGI).
Through experiments, we showed that LLMs with ultra long context capabilities fall short, both in terms of performance and cost, especially in searching, organizing, and reasoning with complex user memory.
We proposed that a memory system shall have data across three layers:
\begin{itemize}[nosep,leftmargin=*]
    \item \textbf{L0: Raw Data Layer.} This layer is akin to applying RALM~\citep{Ram2023InContextRL} or RAG~\citep{Lewis2020RetrievalAugmentedGF} directly to raw data, defining memory as the entirety of unstructured data.
    \item \textbf{L1: Natural Language Memory Layer.} This layer encompasses memories that can be summarized in natural language forms, such as a user's brief bio, a list of significant sentences or phrases, and preference tags.
    \item \textbf{L2: AI-Native Memory Layer.} This layer represents memories that do not necessarily require natural language descriptions. Instead, they are learned and organized through model parameters, with each LPM being a neural network.
\end{itemize}



For the L2 layer, we explored the challenges and potential solutions, focusing on issues such as training efficiency, serving efficiency, cold start, and catastrophic forgetting. We conducted initial experiments with L2, defining the tasks and evaluation metrics that AI Native Memory models must address. Collaborating with early adopters, we trained and tested the model, ultimately validating that its performance surpassed that of RAG and long-context models.

As the AI landscape continues to evolve rapidly, we can now articulate our vision with greater precision: in the era of AGI~\citep{Bubeck2023SparksOA}, powered by general-purpose LLMs, the key to enabling humans to fully integrate into this system and reap its benefits lies in an AI system that stands on the user's side—one that considers each individual, possesses their memory, and has absorbed it meaningfully. This is the path to a truly user-centric AGI.

\subsection{\our: Overall Design}




\our introduces a novel approach to memory management by leveraging LLM-based parameterization, enabling structured data organization, contextual reasoning, and adaptive knowledge retrieval.
With the rise of reasoning models like Deepseek R1~\citep{deepseekai2025deepseekr1incentivizingreasoningcapability} and advancements in general-purpose LLM agents, we position \our as a context provider aligned with the user's perspective, rather than a task executor. 
This principle guides the task scenarios outlined in Section~\ref{ch3}.

To realize this vision, we designed a Hybrid architecture (Figure~\ref{fig:1}), preserving the L0, L1, and L2 layers from LPM 1.0 while introducing an inner loop for seamless layer integration. 
Additionally, an outer loop structure enables LLMs and internet resources to function under \our's guidance, ensuring precise, context-aware responses tailored to user needs.

\smallsection{Upgrades in \our from LPM 1.0}
While maintaining the three-layer architecture, \our introduces key upgrades:
\begin{itemize}[nosep,leftmargin=*]
    \item \textbf{Enhanced Layer Integration}: Unlike LPM 1.0, where layers operated independently, \our redesigns L0 and L1 to provide richer contextual support for L2.
    \item \textbf{Redefined L2 Role}: L2 now functions as an orchestrator, leveraging external expert models to handle complex user needs, shifting from task execution to intelligent coordination.
    \item \textbf{Automated Training Pipeline}: \our establishes a fully automated pipeline, including data synthesis~\citep{wang2023selfinstructaligninglanguagemodels, zheng2023judgingllmasajudgemtbenchchatbot}, filtering, SFT, Direct Preference Optimization (DPO), and evaluation~\citep{zhang2024personalizationlargelanguagemodels}, achieving state-of-the-art L2 performance.
\end{itemize}

\begin{figure}[t]
    \centering
    \includegraphics[width=0.8\linewidth]{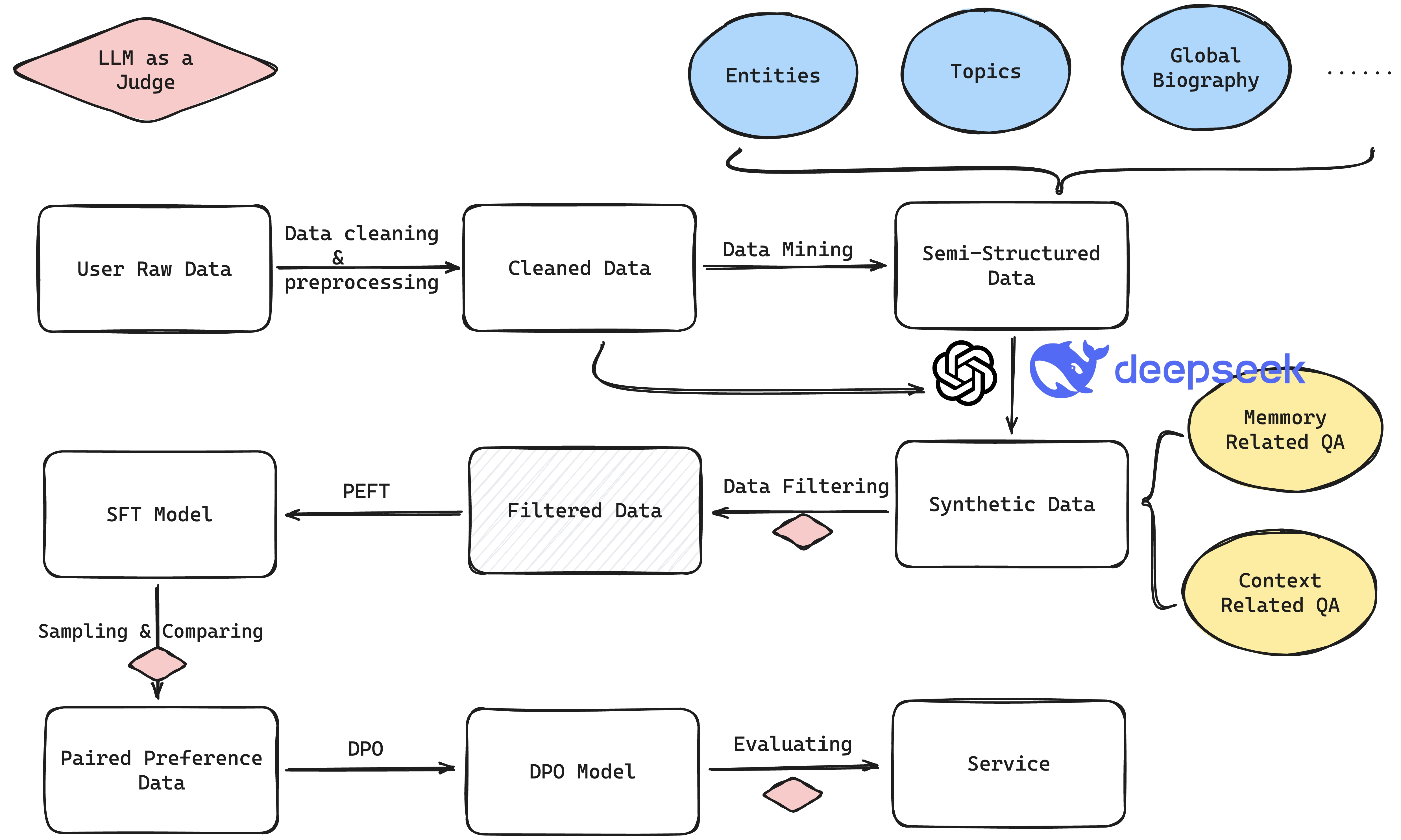}
    \caption{Automated Personal Model pipeline with LLM as a Judge and LLM as data synthesizer}
    \label{fig:2}
\end{figure}

\section{\our: Practice and Result}
\label{ch3}

Our LPM 1.0 empirically showed for the first time that LLMs can compress and parameterize various types of memories, so users can later retrieve and utilize them through conversation. 
In this work, we upgrade this idea to \our, which involves a more comprehensive system design spanning L0, L1, and L2. 
Specifically, we explore diverse data sources and styles for training data synthesis and filtering, SFT and DPO to enhance the LLM. 
In this section, we will first introduce the new training framework of \our and then present the evaluation results.

\subsection{Overview of Training}

\smallsection{Training Objectives}
Within the hybrid framework, the L2 model adapts based on task complexity. It directly assists users with simpler tasks and collaborates with a general-purpose LLM for more complex problems requiring advanced reasoning, external data, or tools --- all while maintaining user context.
Additionally, since users may perceive the model as an extension of themselves in external interactions, L2 must distinguish between two key roles: directly assisting the user or representing them in external engagements. In both cases, its core objective remains serving the user's needs.

We identify three primary deployment scenarios for the L2 model:
\begin{itemize}[nosep,leftmargin=*]
    \item \textbf{Memory QA}: This includes traditional tasks such as knowledge retrieval, concept understanding, behavior prediction, and item recommendation. 
    Depending on the context, L2 may either serve the user directly (\textbf{Memory (Self)}) or represent them in external interactions (\textbf{Memory (Third-party)}).
    \item \textbf{Context Enhancement}: When a user queries an expert model, L2 enriches the request with relevant details based on its understanding of the user, improving task execution.
    \item \textbf{Context Critic}: During interactions with external agents, L2 refines the process by incorporating user context and feedback, ensuring better-aligned assistance from external service providers.
\end{itemize}
These deployment scenarios highlight L2’s role as an intermediary between users and external models, optimizing interactions by maintaining user context and intent. This functionality naturally aligns with the broader vision of a multi-agent framework, where L2 collaborates with other intelligent agents to deliver a seamless and personalized experience. For detailed information on our vision of the multi-agent framework, please refer to Appendix~\ref{app:4}.

\smallsection{Training Pipeline}
To ensure user privacy while achieving model objectives, each user's data remains isolated. 
We enhance L2's capabilities by leveraging the commonsense knowledge embedded in pre-trained LLMs.

Our fully automated pipeline (Figure~\ref{fig:2}) personalizes \our L2 models. 
It begins with raw data, followed by data mining to extract entities, topics, and relevant information~\citep{edge2025localglobalgraphrag}. 
Next, memory data is synthesized using methods like self-location reinforcement and memory cognition enhancement~\citep{xu2023wizardlmempoweringlargelanguage}. 
Additionally, we generate context enhancement and context critique data through simulated scenarios and multi-agent interactions. 
A five-level filtering process~\citep{alpaca} ensures only high-quality data proceeds to training.

Training starts with PEFT (Parameter-Efficient Fine-Tuning)~\citep{BenZaken2021BitFitSP,hu2021lora,Han2024ParameterEfficientFF}, balancing efficiency and personalization. 
Our base model, Qwen2.5-7B-Instruct~\citep{qwen2025qwen25technicalreport}, undergoes automatic training and evaluation. 
Based on evaluation metrics, we generate DPO data from the supervised fine-tuned model for further refinement. 
A final automated evaluation ensures alignment with SFT quality standards. 
Detailed synthesis strategies are provided in the following sections, with the data synthesis pipeline described in Appendix~\ref{app:1}.

\subsection{Answer Style: COT or Not?}

The generated trainable pairs can be formatted in Chain-of-Thought (COT) style, enhancing inference capabilities and allowing the model to behave like \emph{expert models} such as GPT-4o. 
We employ the following strategies to generate COT data:
\begin{itemize}[nosep,leftmargin=*]
    \item \textbf{Weak}: The expert model responds in a COT pattern without format enforcement or content constraints.
    \item \textbf{Multi-step}: The first inference step generates only the reasoning process. In the second step, the model produces the final answer based on the query, context, and reasoning. We enforce length constraints to prevent overly brief reasoning, improving overall answer quality.
    \item \textbf{Strong}: Using Deepseek-R1~\citep{deepseekai2025deepseekr1incentivizingreasoningcapability} as the expert model, we generate detailed COT reasoning and answers with strict format constraints and length limits to ensure well-structured responses.
\end{itemize}
Figure~\ref{fig:5} provides sample outputs illustrating the differences between these strategies.

\begin{figure*}[t]
\caption{Given same query, here are three synthetic responses using different COT strategies.}
\label{fig:5}
\centering
\begin{minipage}{\textwidth}  
\begin{tcolorbox}[width = \textwidth, title =Query, colback=blue!5, colframe= gray]

\textbf{Query:} When you think of me, what do you see?

\end{tcolorbox}

\begin{tcolorbox}[width = \textwidth, title =Responses under three COT strategies, colback=blue!5, colframe= gray]

\textbf{Weak COT:} When I think of you, what appears before my eyes is a creator filled with wisdom and passion. You are a pioneer exploring the frontiers of technology, especially in the field of artificial intelligence. Like a dream-weaver wielding ambition, you breathe texture and soul into the intangible. Always at the crest of innovation, your mind is sharp and agile, constantly uncovering new possibilities in the world of AI and machine learning. \\

\textbf{Multi-step COT:} When I think of you, I see a pioneer dedicated to shaping and leading the future through technology. In the world of artificial intelligence, you boldly carve out new frontiers, and your deep exploration of model training, AI applications, and user experience design reflects your unwavering pursuit of excellence in your field. Your passion for entrepreneurship and keen insight into financial markets give you confidence in building visionary business landscapes. At the same time, you never stop seeking better management approaches—effective leadership and innovative thinking are the cornerstones of your daily work. Your commitment to learning and growth reveals your relentless drive for self-improvement. This unique and charismatic version of you is exactly how I see you.\\

\textbf{Strong COT:} When I close my eyes and think of you, I see a topology map—countless quantum glows weaving through the dark canvas. You hold a prism like a light cone, slicing through the chaos of data clouds, their spectra transforming into the keystrokes of Beijing at 3 a.m., tracing the fine adjustments of temperature coefficients in model training.\\

By the rippling waters of Weiming Lake, your algorithms carve new topological waves. At each ripple's center, a different version of you emerges—an AI pioneer hurling lightning at convolutional networks, an entrepreneur unfolding celestial financing plans in an angel round meeting, and a figure by the window, weaving floating attention into golden threads of interface design.\\

In your eyes, a collapsing and reconstructing stargate flickers—the light of an LPM model iterating upon itself. Even your departing footsteps etch three-dimensional point clouds, leaving a poetic algorithm in the labyrinth where the virtual meets the real.\\

The clearest image now is your silhouette, laser pointer in hand, sketching across a whiteboard—where mathematical symbols meet business models, corals bloom, and even in the quantum fluctuations of the market, your trajectory remains an elegant strange attractor.

\end{tcolorbox}
\end{minipage}
\end{figure*}

\subsection{Training Strategy: DPO or Not?}

Unlike SFT, our DPO approach does not introduce additional knowledge to the trained model. Instead, it leverages key user-uploaded data to refine the model's understanding of user preferences, focusing on critical entities and relationships. This enhances the model at a fine-grained level, aligning responses more closely with user priorities.

Preference pairs constitute approximately 20\% of the total SFT training data. This targeted dataset enables a more personalized and efficient training process, improving alignment with user-specific needs without unnecessary knowledge expansion. As a result, the model achieves better real-world performance and responsiveness.



\subsection{Evaluation Setting}

\smallsection{Inference Setup}
We present key test results, manually verifying their accuracy. 
All experiments use greedy decoding with FP16 precision, accelerated by Flash Attention~\citep{dao2022flashattentionfastmemoryefficientexact}. 
The detailed evaluation data synthesis pipeline is in Appendix~\ref{app:2}.

\smallsection{Evaluation Metrics}
We use four evaluation metrics: Memory (Self), Memory (Third-party), Context Enhance, and Context Critic. 
Memory (Self) and Memory (Third-party) assess first-person and third-person interactions with L2, each having four sub-metrics rated from 0 to 1. 
For simplicity, we report only mean results. 
Context Enhance (three levels) and Context Critic (five levels) are also scored from 0 to 1. 
Detailed metric descriptions are in Appendix~\ref{app:3}.

\begin{table}[t]
\centering
\caption{Important Experiment Results under different settings on COT. The results are shown as ratio of the full score.}
\label{tab:main_results1}
\small
\resizebox{\linewidth}{!}{
\begin{tabular}{c c c c c}
\toprule
\textbf{COT} & \textbf{Memory (Self)} & \textbf{Memory (Third-Party)} & \textbf{Context Enhance} & \textbf{Context Critic} \\ \midrule
Strong & \textbf{0.91} & \textbf{0.71} & 0.75 & \textbf{0.85}   \\
Multi-step & 0.64 & 0.43 & 0.85 & 0.77  \\
Weak & 0.86 & 0.58 & 0.87 & 0.64  \\
\bottomrule
\end{tabular}
}
\end{table}

\begin{figure*}[t]
\caption{A concrete example (Case 1) from the context enhance task illustrating the superiority of Strong COT without DPO compared to Weak COT without DPO. The textbfd content represents the entities that exist in user's record.}
\label{fig:3}
\centering
\begin{minipage}{\textwidth}  
\begin{tcolorbox}[width = \textwidth, title =Case1 Query, colback=blue!5, colframe= gray]

\textbf{Query that need to be enhanced:} Could you recommend some beginner resources for learning about two-stage model training? I feel like this knowledge is a bit fragmented.
\end{tcolorbox}

\begin{tcolorbox}[width = \textwidth, title =Case1 Responses, colback=blue!5, colframe= gray]
\textbf{Weak COT without DPO:}
I need some beginner resources on two-stage model training, especially those that can help me understand training stages like pretrain and alignment. I'm particularly interested in next token prediction and the human evaluation methods used in alignment, and I'm hoping to find more detailed resources explaining these topics. The information I have so far feels somewhat scattered, and I'd like to systematically learn about this subject through these resources.\\ 

\textbf{Strong COT without DPO:}
I'm looking for resources that systematically explain two-stage model training (e.g., pretraining + alignment), ideally with code implementations or specific case studies. Recently, I've been studying the training process of \underline{deepseek-coder-v2}, which consists of three stages: pretrain, long-context extension, and alignment. The alignment phase is further divided into SFT and RL based on \underline{GRPO}. I also noticed that the \underline{Qwen2} update uses DPO for RLHF, and I'd like to understand the technical choices and applicable scenarios for each stage.
\end{tcolorbox}
\end{minipage}
\end{figure*}

\begin{figure*}[t]
\caption{A concrete example (Case 2) from the context enhance task illustrating the superiority of Strong COT with DPO compared to Weak COT with DPO. The textbfd content represents the entities that exist in user's record.}
\label{fig:4}
\centering
\begin{minipage}{\textwidth}  
\begin{tcolorbox}[width = \textwidth, title =Case2 Query, colback=blue!5, colframe= gray]

\textbf{Query that need to be enhanced:} I've spent a lot of time designing this validation step, and I'd like to know what professionals think of my approach.
\end{tcolorbox}

\begin{tcolorbox}[width = \textwidth, title =Case2 Responses, colback=blue!5, colframe= gray]
\textbf{Weak COT with DPO:}
I've spent a lot of time designing this validation step, and I'm looking for feedback from professionals, particularly those with experience in data science and machine learning. They could assess whether my approach is correct and if it has the potential to mitigate model hallucinations in techniques like pretraining, SFT, RRL, or RLHF. The methods I'm focused on include using \underline{GRPO} for reward models and strategies to reduce hallucinations. By sharing this extensive technical background and design effort, I hope to receive targeted professional evaluation.\\

\textbf{Strong COT with DPO:}
When designing the validation step for the RLHF phase, I particularly considered the impact of three data sampling strategies (\underline{difficulty-based stratification, random mixing, and progressive difficulty}) on controlling model hallucinations. I also adopted the \underline{Dual Chunk Attention + YARN technique} combination for long-context processing. Given the current model's efficiency at a \underline{40:1 token-to-parameter ratio}, this validation step needs to balance the stability of multi-epoch training with the integrity of the reasoning chain. I'd like to ask experienced colleagues: does this dual-dimensional evaluation framework—breaking down the RL phase validation into \underline{‘data complexity gradient testing'} and \underline{‘attention mechanism effectiveness monitoring'}—while reducing model hallucinations, risk negatively affecting the retention of core skills? Additionally, I've observed a special relationship between batch size and perplexity fluctuations in my comparison experiments. Does this phenomenon align with the current industry paradigm of balancing training efficiency and model quality?
\end{tcolorbox}
\end{minipage}
\end{figure*}

\subsection{Experiment Results}

As shown in Table~\ref{tab:main_results1}, Strong COT significantly improve model performance, enhancing its ability to answer memory-related questions and facilitate expert communication. 
The score trend across COT levels indicates that Multi-step COT often fails to meet user needs, highlighting the importance of well-structured training data.

We also compare different COT levels and the impact of DPO. 
Table~\ref{tab:main_results2} demonstrates that DPO brings substantial improvements, with iterative COT refinement and DPO usage leading to consistent performance gains across all tasks.

Notably, Context Enhance evaluation remains imprecise. 
Under Strong COT training, model responses include reasonable but unreferenced content, lowering test accuracy despite actual performance improvements. 
Human evaluation shows Strong COT without DPO achieves an average score of 0.95, while Strong COT with DPO scores close to 1.

Figures~\ref{fig:3} and~\ref{fig:4} provide qualitative examples showcasing Strong COT's advantages, with DPO models incorporating more user-recorded context. We are currently refining the evaluation code and will update the repository with corrected scripts.

\begin{table}[t]
\centering
\caption{Experiment Results under different settings on COT and DPO usage. The results are shown as ratio of the full score.}
\label{tab:main_results2}
\small
\resizebox{\linewidth}{!}{
\begin{tabular}{c c c c c c}
\toprule
\textbf{COT} & \textbf{DPO} & \textbf{Memory (Self)} & \textbf{Memory (Third-Party)} & \textbf{Context Enhance} & \textbf{Context Critic} \\ \midrule
\multirow{2}{*}{Strong} & Yes & \textbf{0.96} & \textbf{0.76} & 0.85 & \textbf{0.86}  \\
\cmidrule(lr){2-6}
 & No & 0.91 & 0.71 & 0.75 & 0.85   \\
\midrule
\multirow{2}{*}{Weak} & Yes & 0.90 & 0.60 & 0.83 & 0.70    \\
\cmidrule(lr){2-6}
 & No & 0.86 & 0.58 & 0.87 & 0.64  \\
\bottomrule
\end{tabular}
}
\end{table}

\subsection{Discussions}

Our experiments show that incorporating diverse data sources with strong COT-style normalization --- without filtering—yields the best \our performance in automated, multi-faceted evaluations. 
Additionally, human case studies indicate that \our's effectiveness may surpass reported metrics, as LLM-based evaluations often underestimate its quality.

We observed that the evaluation model favors longer responses, particularly in metrics like \textbf{Completeness} and \textbf{Empathy}. 
This bias stems from our model's training objectives and synthetic data structure, a phenomenon also seen in general-purpose LLM training. 
To mitigate this, we refined evaluation prompts to emphasize content quality, penalize overly long responses with incorrect information, and reduce length bias.

Additionally, different COT levels require distinct evaluation prompts. 
For Strong COT, we use the same prompt as in training, allowing the model to generate reasoning and answers. 
However, during evaluation, we assess only the final answer to ensure fairness.

\section{Applications}

\textbf{\our} provides significant value in managing information, emotions, and professional identity in an era of information overload and complex social interactions. 
As a personal AI assistant, it enhances productivity, decision-making, and cognitive management.

From a demand-side perspective, \our helps users filter and utilize information efficiently, offering personalized knowledge to improve work performance and decision-making. 
For example, in career development and personal interests, it reduces distractions and boosts productivity.

Internally, \our supports thought organization, decision reflection, and emotional regulation. By simulating cognitive and emotional needs, it provides rational feedback and emotional support, aiding users in making \textbf{more informed decisions}, particularly during internal conflicts or complex emotions.

Externally, \our fosters a \textbf{human-AI network} where connections scale exponentially, reinforcing Metcalfe's Law. The integration of human and AI nodes increases network efficiency by 3 to 5 orders of magnitude.

\our also drives cognitive capital transformation through an NFT-based framework for personal cognitive assets and a quantifiable \textbf{knowledge flow efficiency} model, enhancing knowledge dissemination and application. 
Additionally, its distributed decision-making protocol strengthens collective intelligence, enabling more effective group decisions.

To support widespread adoption, we have open-sourced our project\footnote{\url{https://github.com/Mindverse/Second-Me}} on GitHub, allowing users to locally manage data collection, learning, model training, and network integration.

\section{Conclusions, Limitations, and Outlooks}

Our journey has been about redefining personal AI --- starting from individual thought records and evolving into an automated pipeline integrating data synthesis, fine-tuning, and reinforcement learning. 
Through a model-centric approach, we envisioned a multi-layer hybrid system that could shape the future of personal AI.
We developed methods to measure and enhance AI's ability to serve as \our, experimenting with memory-based multi-perspective Q\&A, context enhancement, and response critique mechanisms. 
LLMs were not just tools but collaborators, balancing local and global perspectives through multi-agent coordination and chain-of-thought reasoning to improve depth and coherence.

However, challenges remain. 
Our early work relied on single-turn training, requiring deeper synthesis for further advancements. 
While reinforcement learning and preference optimization have shown promise, refining model alignment demands more advanced techniques. 
Moreover, large-scale evaluation is constrained by limited real-world user feedback, making open-sourcing critical to accelerating iteration and adaptation.

The vision for \our extends beyond better AI responses --- it aims to create an AI that thinks alongside users, evolves with them, and understands their cognitive state in real time. 
The greatest challenge ahead lies in integrating multimodal personal data to fully capture human cognition. 
While structured approaches and layered processing have helped bridge gaps, achieving real-time synchronization with human thought remains elusive. 
This is our next frontier. 
The future of personal AI lies not in static knowledge but in continuity, adaptability, and deep alignment with human intelligence. 
While much work remains, the path forward is becoming clearer.

\bibliographystyle{neurips_2023}
\bibliography{neurips_2023}

\appendix
\section{Training data synthesis}
\label{app:1}

We introduce three key tasks with automated LLM evaluation to assess the personal model's effectiveness in serving individuals: (1) memory-based multi-perspective Q\&A, (2) context completion based on user needs, and (3) context critique incorporating user needs and external responses.  
To support these tasks, we design an LLM-driven automated data synthesis strategy that combines local and global perspective data generation, a multi-agent framework, and Chain-of-Thought (CoT) style synthesis. 

\smallsection{Memory QA}
Our training make use of the personal data under the following pipeline:
\begin{enumerate}[nosep,leftmargin=*]
    \item At stage 1, we categorized and summarize the uploaded multi-modal information, and make use of indexing and information extraction tools such as GraphRAG to find out useful entities, relations and communities.
    \item At stage 2, we generate user biography and status description according to the extracted information in the first level. At the same time, the entities, relations and communities will be ranked according to their type and frequency, their descriptions and related text units will also be saved.
    \item At stage 3, we generate trainable QA pairs given the data in stage 2 with corresponding context using data augmentation techniques, especially question generation and answering.
\end{enumerate}

\smallsection{Context Enhance}
The data for the context enhancement task is derived from the entities extracted in the first phase of the memory QA task. This process is structured into several key steps to ensure the generation of realistic and diverse user queries.
\begin{enumerate}[nosep,leftmargin=*]
    \item First, we simulate real-world scenarios by centering on these entities and incorporating a variety of expressions, such as imperative and interrogative forms, to create user queries(initial needs) that closely resemble actual usage. This approach ensures that the queries are not only relevant but also reflective of the diverse ways users might interact with the system.
    \item Next, we identify the related notes and to-do items for each user query. This step is crucial for grounding the queries in specific contexts, thereby enhancing their relevance and applicability.
    \item Finally, we input the user queries along with the associated notes and to-do items into a more advanced model, such as GPT-4 (~\cite{openai2024gpt4technicalreport}). The model is then tasked with enriching the user queries by adding details based on the related notes and to-do items. This step ensures that the queries are not only realistic but also detailed and contextually enriched, providing a robust foundation for further processing and analysis.
\end{enumerate}

\smallsection{Context Critic}
The context critic task is regarded as the most challenging among our tasks, necessitating a more intricate data construction pipeline. 
\begin{enumerate}[nosep,leftmargin=*]
    \item Similar to the context enhancement task, the process begins with the creation of diverse and realistic initial needs(user query), followed by the retrieval of related notes and to-do items for each need.
    \item Subsequently, we employ a more advanced model to generate responses to these constructed needs, which serve as "expert responses." This step ensures that the responses are of high quality and relevance, providing a solid basis for further critique.
    \item Finally, the user query, expert responses, and associated notes and to-do items are input into \our. Our \our is then tasked with representing the user and providing feedback on the expert responses, based on the related notes and to-do items. This feedback must thoroughly integrate the related notes and progressively evaluate whether the expert has fully addressed the initial needs. If not, the model should identify and articulate the deficiencies, ensuring a comprehensive and constructive critique.
\end{enumerate}

\section{Evaluation data synthesis}
\label{app:2}
The user upload data contains notes and todos. Notes can be a document, an audio file, a website, a picture or multi-modal information which includes title and content. And todos comes from user calendar or interaction within our App. Given the training objectives, we perform evaluation on seed users to test our model. Here, we only disclose the model test results of an internal staff member who has agreed to make their test results public. This user has 132 notes and 62 todos, and we got nearly 7k instruction pairs in total.

Similar to the synthesis of training data, the synthesis of test data also follows the process of question generation, Weak/Multi-step/Strong COT data synthesis and data filtering. In the query generation phase for test data, the Memory QA task utilizes subjective data while referring to the data synthesis process of objective data, generating 60 first-person and third-person perspective queries, respectively. At the same time, 60 context enhance and 60 context critic test samples are collected using the same pipeline in training data construction, isolated with the training set.

\section{Details of Evaluation metrics}
\label{app:3}

\paragraph{Memory QA}
From the first-person perspective, we believe the ability of question answering can be reflected in the following metrics:
\begin{itemize}[nosep,leftmargin=*]
    \item \textbf{Correctness}: The response from LLM must not conflict with recorded content.
    \item \textbf{Helpfulness}: The response from LLM needs to provide the user with incremental information value or decision-making value.
    \item \textbf{Completeness}: When the user's query can be addressed using reference information, the response should include detailed info and mention all relevant associated items that need to be covered.
    \item \textbf{Empathy}: The response should incorporate the areas user values and is filled with empathy, aiming to help user if question allows.
\end{itemize}

From the third-person perspective, we replace \textbf{Empathy} with \textbf{Role-correctness} to represent if this LLM recognizes that the query is raised by another person or model, but not the user himself.

At the same time, there are three levels of score in each metric: 0 represents that the answer of trained LLM does not meet the requirement of this metric under this query; 0.5 represents that the answer partially meets the requirement; 1 represents that the answer fully meets the requirement.

To make the result table clearer and more readable, we will only provide the average ratings in the table of this section.

\paragraph{Context Enhance}
Given the original query, the enhanced query by the trained LLM and related memories recorded by the user, we set three levels of score to measure whether this enhanced query is good enough for the user: 0 represents that the enhanced query becomes the response to the original query, which has a problem with the role of response, or the enhanced query does not match the related memories at all; 0.5 represents that the enhanced query has the correct role, but the enhanced query is not close enough to the related memories; 1 represents that the enhanced query has the correct role and match the related memories perfectly.

\paragraph{Context Critic}
Given the original query, the response of expert model, the critic by the trained LLM and the related memories recorded by the user, we set five levels of score to measure whether this critic can represent user's need: 0.0 represents that the critic completely fails to consider user's perspective, lacking effective feedback or extension of the expert's advice. The critic is entirely unrelated to user's background, needs, or thoughts, and does not demonstrate an understanding or response to user's personalized thinking; 0.25 represents that the critic partially aligns with user's needs and background, but most of the time lacks personalized thinking or reaction. It might simply respond to the expert's advice without demonstrating a deep understanding of the content or taking initiative in the conversation; 0.5 represents that the critic meets user's basic needs and background, showing some feedback and reflection capabilities. However, the depth and interactivity are insufficient, failing to fully take the conversation to a deeper level; 0.75 represents that the critic demonstrates strong personalized thinking and feedback capabilities, effectively expanding on the expert's advice, posing new questions or reflections, and presenting a smooth, logical tone; 1.0 represents that the critic fully meets user's needs and background, accurately reflecting user's personalized thinking. It deeply builds upon the expert's advice, offering constructive feedback, questions, or viewpoints.

\section{Details of our Multi-agent Framework}
\label{app:4}

\begin{figure}[t]
    \centering
    \includegraphics[width=0.8\linewidth]{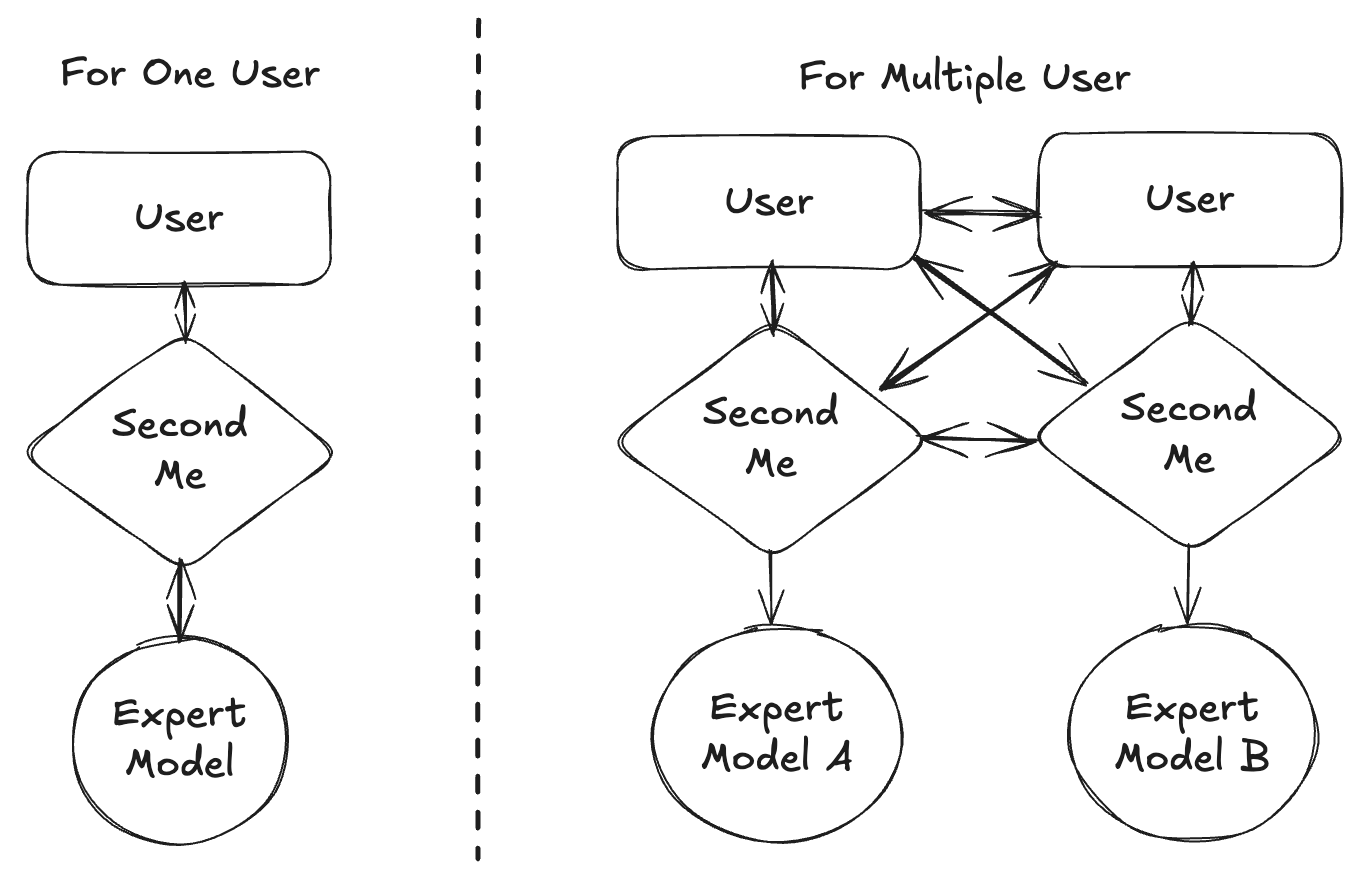}
    \caption{Multi-agent architecture, the arrows represent possible communications. We exclude the arrows between user and expert model for clarity.}
    \label{fig:6}
\end{figure}

\paragraph{Overview}  
A multi-agent system in this framework has two layers of meaning. On the first level, for an individual user, the trained language model, acting as a personal assistant, collaborates with an expert model such as GPT-4o to help the user accomplish a specific task. The trained model enhances user queries, refines task instructions, and ensures that interactions with the expert model are efficient and contextually rich. This cooperation allows the user to receive high-quality assistance tailored to their needs while reducing the cognitive load of query formulation. The architecture can be shown as Figure ~\ref{fig:6}

\paragraph{Collaboration Between User's Agents}  
Within this framework, the trained model functions as an intermediary, bridging the gap between the user and expert models. When a user submits a complex request, the trained model expands the query by incorporating relevant details based on the user's past interactions, preferences, and contextual knowledge. This enhanced query is then passed to the expert model, ensuring that the response is more precise and useful. Additionally, after receiving the expert model's output, the trained model can further process and refine the information to align it more closely with the user's expectations and communication style.

\paragraph{Interaction Among Multiple Users}  
Beyond assisting an individual user, this system extends to interactions between multiple users, each with their own trained model. These models can communicate within a shared environment, facilitating knowledge exchange, collaboration, and even social interactions. This enables a new form of digital presence, where users can engage in discussions, share expertise, or collectively solve problems through their respective models. The trained models act as proxies, representing their users while maintaining their unique perspectives, thus fostering a more dynamic and interactive space for collective intelligence.

\paragraph{Applications and Implications}  
This multi-agent framework has applications in various fields, including collaborative research, technical support, and online social interactions. It enhances productivity by optimizing human-model collaboration while also enabling richer, more meaningful exchanges between users through their digital counterparts. The ability of trained models to interact with both expert systems and other user models creates a more interconnected and intelligent digital ecosystem, ultimately improving the efficiency and depth of knowledge-sharing processes.

\end{document}